\documentclass[runningheads]{llncs}
\usepackage{graphicx}
\usepackage{comment}
\usepackage{amssymb} % define this before the line numbering.

\usepackage{multicol}
\usepackage{multirow}
\usepackage{tikz}
\usepackage{adjustbox}
\usepackage{algorithm}               % format of the algorithm
\usepackage{algorithmic}             % format of the algorithm
\usepackage{booktabs}
\usepackage{epstopdf}
\usepackage{enumerate}
\usepackage{pdfpages}
\usepackage{afterpage}
\usepackage{color, colortbl}
\usepackage{footmisc}
\usepackage{etoolbox}
\usepackage{tablefootnote}
\usepackage{amsmath}
\usepackage{url}
\usepackage{graphicx}

\usepackage{subfig}
\usepackage{caption}
\usepackage[flushleft]{threeparttable}

\usepackage[accsupp]{axessibility}  % Improves PDF readability for those with disabilities.
\graphicspath{{./figures/small/}}

\begin{document}
% \linenumbers
\pagestyle{headings}
\mainmatter
\def\ECCVSubNumber{3794}  % Insert your submission number here

\title{EASNet: Searching Elastic and Accurate Network Architecture for Stereo Matching} % Replace with your title

\begin{comment}
\titlerunning{ECCV-22 submission ID \ECCVSubNumber} 
\authorrunning{ECCV-22 submission ID \ECCVSubNumber} 
\author{Anonymous ECCV submission}
\institute{Paper ID \ECCVSubNumber}
\end{comment}
%******************

%\begin{comment}
\titlerunning{EASNet}
%\author{First Author\inst{1}\orcidID{0000-1111-2222-3333} \and
%Second Author\inst{2,3}\orcidID{1111-2222-3333-4444} \and
%Third Author\inst{3}\orcidID{2222--3333-4444-5555}}
%
\author{Qiang Wang\inst{1,2,3} \and
Shaohuai Shi\inst{4} \and
Kaiyong Zhao\inst{5,3} \and
Xiaowen Chu\inst{6,4,3}\thanks{Corresponding author.}}
\authorrunning{Q. Wang et al.}
\institute{
Harbin Institute of Technology (Shenzhen), Shenzhen, China\\
\email{qiang.wang@hit.edu.cn}\\
\and
Guangdong Provincial Key Laboratory of Novel Security Intelligence Technologies, Shenzhen, China
\and
Hong Kong Baptist University, Hong Kong SAR, China\\
\email{\{qiangwang,kyzhao,chxw\}@comp.hkbu.edu.hk}\\
\and
The Hong Kong University of Science and Technology, Hong Kong SAR, China\\
\email{shaohuais@cse.ust.hk}\\
\and
XGRIDS, xgrids.com, Shenzhen, China\\
\email{kyzhao@xgrids.com}
\and
The Hong Kong University of Science and Technology (Guangzhou), Guangzhou, China\\
\email{xwchu@ust.hk}
}
%\end{comment}
%******************
\maketitle

\begin{abstract}
Recent advanced studies have spent considerable human efforts on optimizing network architectures for stereo matching but hardly achieved both high accuracy and fast inference speed. To ease the workload in network design, neural architecture search (NAS) has been applied with great success to various sparse prediction tasks, such as image classification and object detection.
However, existing NAS studies on the dense prediction task, especially stereo matching, still cannot be efficiently and effectively deployed on devices of different computing capability. 
%Inspired by once-for-all (OFA), which is a novel NAS method for image classification, 
To this end, we propose to train an \underline{e}lastic and \underline{a}ccurate network for \underline{s}tereo matching (EASNet) that supports various 3D architectural settings on devices with different compute capability. Given the deployment latency constraint on the target device, we can quickly extract a sub-network from the full EASNet without additional training while the accuracy of the sub-network can still be maintained.
Extensive experiments show that our EASNet outperforms both state-of-the-art human-designed and NAS-based architectures on Scene Flow and MPI Sintel datasets in terms of model accuracy and inference speed. Particularly, deployed on an inference GPU, EASNet achieves a new SOTA 0.73 EPE on the Scene Flow dataset with 100 ms, which is 4.5$\times$ faster than LEAStereo with a better quality model. The codes of EASNet are available at: \textcolor{blue}{\url{https://github.com/HKBU-HPML/EASNet.git}}
\end{abstract}

\sloppy
\section{Introduction}
Stereo matching, also called disparity estimation, is a conventional but important technique widely applied to various computer vision tasks, such as 3D perception, 3D reconstruction and autonomous driving. Stereo matching aims to find dense correspondences between a pair of rectified stereo images. 
As traditional stereo matching algorithms with manual feature extraction and matching cost aggregation fail on those textureless and repetitive regions in the images due to lack of their prior information, deep neural network (DNN) based methods avoid this failure by efficiently learning the data distribution and have achieved state-of-the-art (SOTA) performance in many public benchmarks \cite{mayer2016large,butler2012naturalistic,kitti2012,kitti2015} in recent years. However, DNN networks for stereo matching should also be well designed to achieve good performance. Existing human-designed stereo networks can be divided into two main classes, the U-shape network with 2D convolution (U-Conv2D) and cost volume aggregation with 3D convolution (CVA-Conv3D).

The U-Conv2D methods leverage the U-shape encoder-decoder structure with 2D convolution layers to directly predict the disparity map. The representative networks are the DispNet/FlowNet series \cite{flownet1,flownet2,flownet3,mayer2016large} as well as their variants \cite{crl2017,wang2020fadnet,wang2021fadnet++}. This category of networks enjoys computing efficiency of 2D convolution. However, recent studies \cite{cheng2020hierarchical} raise some concerns about the generalization ability of the U-Conv2D methods. 
%For example, the DispNet series fails the random dot stereo tests \cite{Zhong_2018_ECCV}. 
In contrast, the CVA-Conv3D methods exploit the concept of semi-global matching \cite{sgm2007} and construct a 4D feature volume by aggregating features from each disparity-shift to enhance the generalization ability. In CVA-Conv3D, it firstly constructs cost volumes by concatenating left feature maps with their corresponding right counterparts across each disparity candidate \cite{gcnet2017,psmnet2018,ganet2019,ednet2021}. The cost volumes are then automatically aggregated and regressed by 3D convolution layers to produce the disparity map. This branch of methods nowadays achieves SOTA estimation quality and dominates the leader-board of several public benchmarks \cite{kitti2012,kitti2015}. However, due to the expensive computing cost of 3D operations, they typically run very slowly and are difficult for real-time deployment even on the modern powerful AI accelerators (e.g., GPUs). 

On the other hand, AutoML \cite{he2019automl} techniques (e.g., neural architecture search (NAS)~\cite{nas2019}) recently become very popular to relieve AI practitioners from manual trial-and-error effort by automating network design. Recent years have witnessed tremendous successes of NAS in various computer vision tasks (e.g., classification \cite{Zoph_2018_CVPR,Real_Aggarwal_Huang_Le_2019}, object detection \cite{efficientdet2020}, and semantic segmentation \cite{Nekrasov_2019_CVPR,Liu_2019_CVPR}). However, existing applications of NAS are mainly used on sparse prediction problems like classification and object detection. It would become very challenging to apply NAS to dense prediction problems like stereo matching because of the following two reasons. 1) In general, NAS needs to search through a humongous set of possible architectures to determine the network components, which requires extensive computational costs (e.g., thousands of GPU hours). 2) The memory footprint and the model computation workload of stereo matching networks are much larger than those of sparse prediction networks. Taking an example of two architectures, GANet \cite{ganet2019} and ResNet-50 \cite{resnet2016}, on stereo matching problems and image classification problems, respectively. To process one sample on an Nvidia Tesla V100 GPU, GANet requires nearly 29 GB of GPU memory and 1.9 seconds inference time (on the Scene Flow \cite{mayer2016large} dataset), while ResNet-50 only requires 1.5 GB and 0.02 seconds (on the ImageNet \cite{deng2009imagenet} dataset). Therefore, directly applying the strategies of sparse prediction in NAS to stereo matching can lead to prohibitive workloads due to the explosion of computational resource demands. 

To avoid such a problem, Saikia et al. \cite{Saikia_2019_ICCV} propose AutoDispNet that searches the architecture based on the U-Conv2D methods, and it limits its search space on three different cell structures rather than the full architecture. Although AutoDispNet saves search time, it still cannot surpass the existing SOTA CVA-Conv3D methods (e.g. AANet \cite{xu2020aanet}) in both model accuracy and inference speed. Later, Cheng \cite{cheng2020hierarchical} leveraged task-specific human knowledge in the search space design to reduce the demands of computational resources in searching architectures, and proposed an end-to-end hierarchical NAS network named LEAStereo, which achieved the SOTA accuracy among the existing CVA-3D methods. However, LEAStereo takes 0.3 seconds of model inference even on a high-end Nvidia Tesla V100 GPU, which is far away from the requirement of real-time applications. Moreover, both AutoDispNet and LEAStereo attempt to find a specialized network architecture, and train it from scratch, thus cannot be scaled to different devices. Notice that the deployment of stereo matching applications can have diverse computing resource constraints, from high-end cloud servers to low-end edge devices or robotics embedded devices. To meet the latency requirement of a new given device, the above two methods need to re-search and re-train the model, which requires large labor. The recent proposed once-for-all (OFA) network \cite{cai2020once} tries to support diverse architectural settings, but it only explores the sparse prediction problem like image classification. In summary, existing stereo matching methods including human-designed architectures and searched architectures cannot well fulfill real-world deployment requirements which need to consider model accuracy, inference speed, and training cost. 
%To this end, we design a solution to address this problem in this paper.
\iffalse
\begin{figure}[!t]
	\centering
	\includegraphics[width=0.76\linewidth]{figures/sf_comparison.pdf}
	\caption{Our proposed method, EASNet, sets a new state-of-the-art on the Scene Flow \textsl{test} dataset with much fewer parameters and much lower inference time. The data points on the EASNet line indicate different sub-networks sampled from its largest full network structure.}
	\label{fig:sf_intro}
\end{figure}
\fi

To this end, we propose to train an \underline{e}lastic and \underline{a}ccurate \underline{s}tereo matching network, EAS-Net, which enables model deployment on devices of different computing capability to guarantee the inference speed without additional training while the model accuracy can be maintained. 
%As illustrated in Fig. \ref{fig:sf_intro}, our EASNet achieves Pareto optimality among all the methods in terms of model accuracy and inference speed.
Furthermore, our EASNet does not need to re-train or re-search the architecture for deployment on any new devices. In this paper, we make three-fold contributions:
% \vspace{-0.5 em}
% \vspace{-0.5pt}
\begin{itemize}
    \item Based on the pipeline of the CVA-3D methods, we propose an end-to-end stereo matching network named EASNet that contains four function modules. We allow the network to search for both the layer level and the network level structures in a huge network candidate space.
    \item To efficiently train EASNet, we develop a multiple-stage training scheme for EASNet to reduce the model size across diverse dimensions of network architecture parameters including depth, width, kernel size and scale. Our training strategy can significantly improve the prediction accuracy of sub-networks sampled from the largest full EASNet structure, which enables flexible deployment according to the target device computing capability and latency requirement without any additional model training. 
    \item We conduct extensive experiments to evaluate the effectiveness of EASNet on several popular stereo datasets among three GPUs of different computing power levels. Under all deployment scenarios, EASNet outperforms both the human-designed and NAS-based networks in terms of model accuracy and inference speed. 
    %Deployed on an inference GPU, EASNet achieves a new SOTA 0.73 EPE on the Scene Flow dataset with 100 ms, which is 4.5$\times$ faster than LEAStereo with a better quality model. 
    % We also achieve competitive performance on Sintel and KITTI datasets to prove the generalization of EASNet.
\end{itemize}
%The rest of the paper is organized as follows. We introduce some related work in deep neural network methods for the stereo matching problem in Section \ref{sec:related_work}. Section \ref{sec:model} introduces the methodology and implementation of our proposed EASNet. We demonstrate our experimental results in Section \ref{sec:exp}. We finally conclude the paper in Section \ref{sec:conclusion}.

\section{Related Work}\label{sec:related_work}
\subsection{Manual DNN Design for Stereo Matching}
% Over the past five years, the stereo matching problem has been well addressed by leveraging deep learning methods to extract effective features from a pair of stereo images and estimate their correspondence cost. The existing methods of manual design can be roughly classified into two categories: the U-shape network with 2D convolution (U-Conv2D) and cost volume aggregation with 3D convolution (CVA-Conv3D).

In recent years, many deep learning methods have been proposed for stereo matching by extracting effective features from a pair of stereo images and estimating their correspondence cost, which can be classified into 2D with the U-shape network (U-Conv2D) and 3D with cost volume aggregation (CVA-Conv3D) methods. 

On the one hand, in the U-Conv2D networks, the U-shape network architecture mainly utilizes 2D convolution layers \cite{mayer2016large}\cite{crl2017} to estimate disparity, which takes a pair of rectified stereo images as input and generates the disparity by direct regression. However, the pure 2D CNN architectures are difficult to capture the matching features such that the estimation results are not good. 
%To address the problem, the correlation layer which can express the relationship between left and right images is introduced in the end-to-end architecture (e.g., DispNetCorr1D \cite{mayer2016large}, FlowNet \cite{flownet1}, FlowNet2 \cite{flownet2}, and DenseMapNet \cite{fast2018}). The correlation layer significantly boosts the estimating performance compared to the pure CNNs. However, recent studies raise some concerns about their generalization. For example, the DispNet \cite{mayer2016large} fails the random dot stereo tests \cite{Zhong_2018_ECCV}.
On the other hand, the 3D methods with cost volume aggregation, named CVA-Conv3D, are further proposed to improve the estimation performance \cite{zbontar2016stereo}\cite{gcnet2017}\cite{psmnet2018}\cite{ganet2019}\cite{nie2019multi}, which apply 3D convolutions to cost volume aggregation. The cost volume is mainly constructed by concatenating left feature maps with their corresponding right counterparts across each disparity level \cite{gcnet2017}\cite{psmnet2018}, and the features of the generated cost volumes can be learned by 3D convolution layers. Nowadays the top-tier CVA-Conv3D methods \cite{ganet2019,xu2020aanet,cheng2020hierarchical} have achieved very good accuracy on various public benchmarks. However, the key limitation of CVA-Conv3D is its high computation resource requirements, which makes them be difficult for real-world deployment. For example, GANet \cite{ganet2019} and LEAStereo \cite{cheng2020hierarchical} take 1.9 seconds and 0.3 seconds respectively on predicting the disparity map of a stereo pair of 960$\times$540 even using a very powerful Nvidia Tesla V100 GPU. Though they achieve good accuracy, it is difficult to deploy them for real-time inference. 

\subsection{NAS-based Stereo Matching} 
%To lessen the effort dedicated to designing network architectures, automated machine learning (AutoML) \cite{he2019automl}, especially neural architecture search (NAS) \cite{nas2019,zoph2016neural,Zoph_2018_CVPR,Real_Aggarwal_Huang_Le_2019,liu2018darts,Cai_Chen_Zhang_Yu_Wang_2018}, has become an increasingly active research area over the past few years. While most of the early studies \cite{zoph2016neural,optimizer2017,cai2017reinforcement,augment2002,liu2017hierarchical,evolve2017} targeted the sparse prediction tasks, such as image classification and object detection. 
%They leveraged reinforcement learning and evolution algorithm as an agent to search the network architecture. Due to the huge search space, the reinforcement learning based methods typically require up to thousands of GPU days, which restricts their applications in many practical scenarios with limited computing resources. 
%To accelerate the search procedure, some works \cite{liu2018darts,oneshot2018} reformulated the NAS problem by training a large super network that contains all the candidate architectures. The representative is DARTS \cite{liu2018darts}, which makes the links between different operation and cell candidates and finalizes the network structure according to the link importance. 
To lessen the effort dedicated to designing network architectures, AutoML \cite{he2019automl}, especially NAS \cite{nas2019,Zoph_2018_CVPR,Real_Aggarwal_Huang_Le_2019,liu2018darts,Cai_Chen_Zhang_Yu_Wang_2018}, has become an increasingly active research area over the past few years. While most of the early studies \cite{optimizer2017,cai2017reinforcement,augment2002,liu2017hierarchical,evolve2017,liu2018darts} have proven the effectiveness of NAS in many sparse prediction tasks, the extension to dense prediction tasks, such as semantic segmentation \cite{Nekrasov_2019_CVPR,chen2019fasterseg} and stereo matching \cite{Saikia_2019_ICCV,cheng2020hierarchical}, is still at an early stage. 
%Compared to the popularity and diversity of NAS approaches for sparse prediction tasks, the extension to dense prediction tasks, such as semantic segmentation \cite{Nekrasov_2019_CVPR,chen2019fasterseg} and stereo matching \cite{Saikia_2019_ICCV,cheng2020hierarchical}, is still at an early stage. 
AutoDispNet is the first work that adopts the DARTS NAS method to search the efficient basic cell structure for the U-Conv2D method. However, to reduce the prohibitive search space, it only searches partially three different cell structures rather than the full architecture methods, and thus achieves limited model accuracy and generalization. In contrast, LEAStereo \cite{cheng2020hierarchical} leveraged the domain knowledge of stereo matching and designed a hierarchy end-to-end pipeline, which allows the network to automatically select the optimal structures.
However, LEAStereo is still difficult to be deployed on modern AI processors due to the high computational cost. Furthermore, to meet the latency requirement on the new target device, LEAStereo needs to tune the network search space accordingly, followed by re-searching and re-training the model, where the expensive network specialization is unavoidable. 

Recent studies \cite{Cai_Chen_Zhang_Yu_Wang_2018,cai2018proxylessnas,Tan_2019_CVPR,jiang2019fpga,hao2019iot,huang2017multi,lin2017runtime} take the hardware capability into account to search the network. As one of the existing SOTA studies, the once-for-all (OFA) network allows direct deployment under various computing devices and constraints by selecting only a part of the original full one without additional model training. 
%The key contribution is the progressive shrinking training strategy that improves the accuracy of all sub-networks sampled from the full OFA network. 
However, they only discuss the case for image classification, while the domain knowledge and processing pipeline of stereo matching are much different from the sparse prediction task of OFA.
In this paper, we propose a novel network named EASNet to search elastic and accurate stereo matching networks, and design a specialized network search space according to the prior geometric knowledge of stereo matching. Our EASNet covers a wide range of search dimensions (kernel size, width, depth, and scale). With negligible accuracy loss and without any extra model fine-tuning, our EASNet can be directly deployed on different scenarios of computing power and resource constraint, which refers to its ``elastic'' and ``accurate'' characteristics.

\section{Our Method: EASNet}\label{sec:model}
In this section, we present our proposed elastic network structure for stereo matching, EASNet, that covers four function modules in the search and training pipeline. We support up to four search dimensions for different modules in EASNet. 
%The OFA methodology \cite{cai2020once} has been proved to efficiently prune the full network architecture of image classification. However, due to the natural task property differences between image classification and stereo matching, we need to extend the progress shrinking solution in \cite{cai2020once} to enable efficient model training of our specialized network search space. 
We firstly describe the architecture search space of each function module, including their basic unit and supported search dimensions. Then we introduce the training approach across four search dimensions to maximizing the average model accuracy of all the derived sub-networks in EASNet.
\subsection{The Architecture Space of EASNet}
\begin{figure*}[!t]
	\centering
	\includegraphics[width=0.88\linewidth]{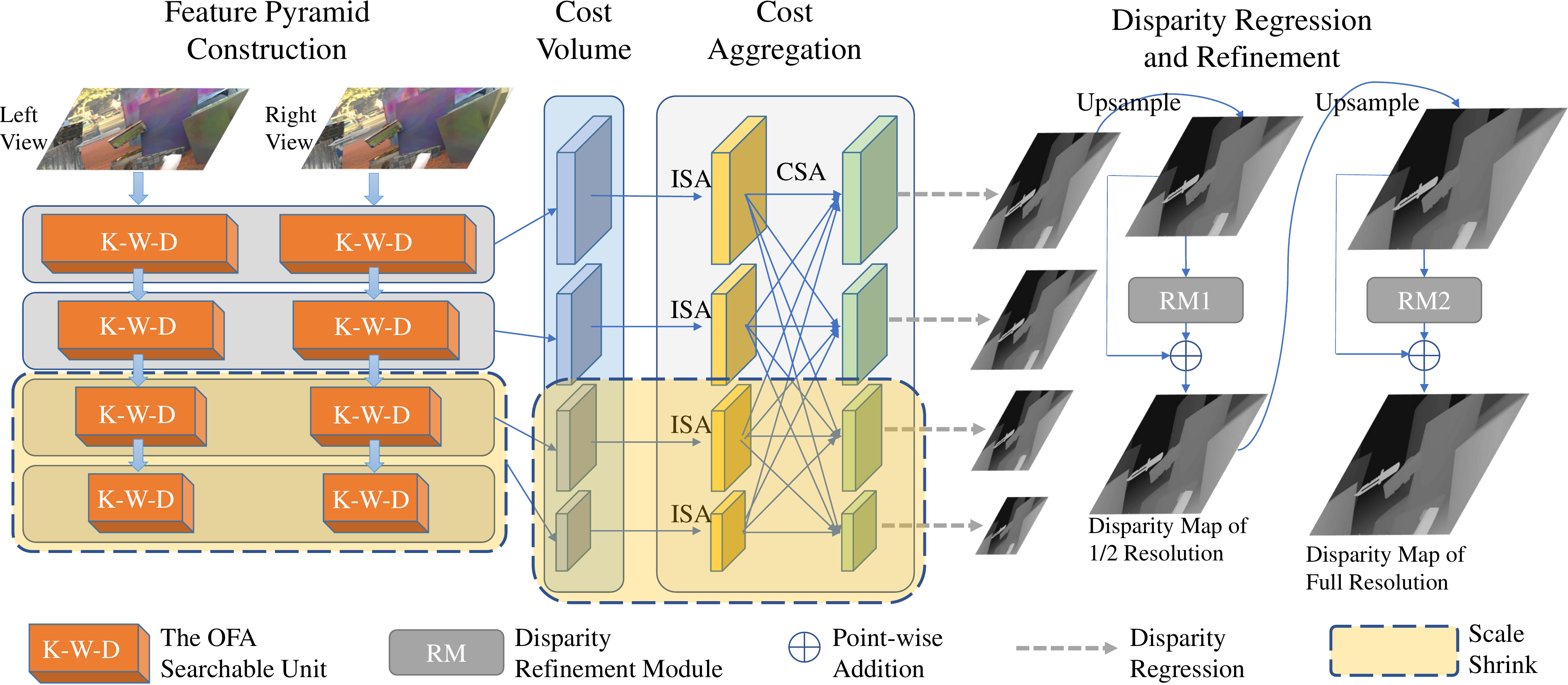}
	\caption{The model structure of our proposed EASNet. It contains four modules with different functions derived from the domain knowledge of stereo matching. The OFA searchable unit applies the similar methodology in \cite{cai2020once}. The parts covered by the shallow yellow dotted blocks can be alternatively skipped when applying scale shrinking. %The part covered by the shallow blue dotted block can be alternatively skipped when applying refinement shrinking.
	}
	\label{fig:easnet}
\end{figure*}

In this subsection, we introduce the overview structure of EASNet. As illustrated in Fig. \ref{fig:easnet}, EASNet is composed of four modules: feature pyramid construction, cost volume, cost aggregation, and disparity regression and refinement. The functions of these modules are benefited from prior human knowledge in stereo matching and success of previous hand-crafted network architecture design. EASNet enables its flexibility and effectiveness by providing different levels of support of neural architecture search for these four modules.

\textbf{Feature Extraction.}
In feature extraction, we need to extract multi-scale features from the input left and right images and construct a feature pyramid for the latter cost volume stage. Thus, we design a sequence of searchable units (similar to~\cite{cai2020once}) that cover three important dimensions of CNNs, i.e., depth, width, and kernel size. The i$^{th}$ unit produces features maps of $1/(3\times2^{i-1})$ resolution by setting stride=2 for the first convolution layer and stride=1 for the rest in it. For example, in our experimental setting, there are totally four units providing different resolutions of feature maps from 1/3 to 1/24. 
We also enable each unit to use arbitrary numbers of layers (denoted as elastic depth) as that of OFA~\cite{cai2020once}. Then we allow each layer to use arbitrary numbers of channels (denoted as elastic width) and arbitrary kernel sizes (denoted as elastic kernel size). In our experimental setting, the depth of each unit is chosen from $\{2,3,4\}$; the width expansion ratio in each layer is chosen from $\{2,4,6,8\}$; the kernel size is chosen from $\{3,5,7\}$.
Therefore, with 4 units, we have roughly $((3\times4)^2 + (3\times4)^3 + (3\times4)^4)^4 \approx 10^{13}$ different architectures. Since all of these sub-networks share the same weights, we only require 5M parameters to store all of them. Without sharing, the total model size will be extremely large, which is impractical.

We further introduce one more dimension of the network search space, the total scale of the feature pyramid (denoted as elastic scale) to EASNet. As proved by existing studies \cite{cheng2020hierarchical}, the number of scales in a feature pyramid can significantly affect the model accuracy of disparity prediction. Deeper feature pyramids typically provide better prediction accuracy but require more computational efforts. Thus, we allow our EASNet to skip some high levels of feature maps and fine-tune the low levels. Take an example shown in Fig. \ref{fig:easnet}, the part covered by the shallow golden dotted blocks can be alternatively skipped during model fine-tune and inference. The scale of feature pyramid is chosen from $\{2,3,4\}$ in our experiments.

\textbf{Cost Volume.} After the feature pyramids of the left and right images are constructed, we then establish the multi-scale 3D cost volume by correlating left and right image features at corresponding scales with the point-wise multiplication operation, which is similar to AANet \cite{xu2020aanet}. 
\begin{align}
	\scriptsize
    \textbf{C}(d,h,w)=\frac{1}{N}\langle\textbf{F}_{l}^{s}(h,w), \textbf{F}_{r}^{s}(h,w-d) \rangle, \label{eq:cal_cv}
\end{align}
where $\langle \cdot,\cdot \rangle$ denotes the inner product of two feature vectors and $N$ is the channel number of extracted features. $\textbf{F}_{l}^{s}$ denotes the feature maps of the scale $s$ extracted from the left view, and $\textbf{F}_{r}^{s}$ refers to the ones from the right view. $\textbf{C}(d,h,w)$ is the matching cost at location $(h,w)$ for disparity candidate $d$. 
Thus, $S$ scales of feature pyramid produce $S$ 3D cost volume. The raw cost volume in this module will be then fed into the cost aggregation module. In the cost volume module, we also support elastic scale which can be chosen from [2,3,4]. The chosen scale is naturally consistent with the scale number of feature pyramid. 

\textbf{Cost Aggregation.}
The cost aggregation module is used to compute and aggregate matching costs from the concatenated cost volumes. We apply the stacked Adaptive Aggregation Modules (AAModules) for flexible and efficient cost aggregation, as it can simultaneously estimate the matching cost in the views of intra-scale and cross-scale. An AAModule consists of $S$ adaptive Intra-Scale Aggregation (ISA) modules and an adaptive Cross-Scale Aggregation (CSA) module for $S$ pyramid levels.

For each scale of the cost volume, ISA can address the popular edge-fattening problem in object boundaries and thin structures by enabling sparse adaptive location sampling. In \cite{xu2020aanet}, ISA is implemented by dilated convolution. In particular, we use the same implementation of ISA in \cite{xu2020aanet}, which is a stack of three layers (i.e., 1$\times$1 convolution, 3$\times$3 deformable convolution and 1$\times$1 convolution) and a residual connection.

Assume that the resulting cost volume after ISA is $\tilde{C}^s$, we apply the CSA module to explore the correspondence among different scales of $\tilde{C}^s$. To estimate the cross-scale cost aggregation of the scale $s$, we adopt
\begin{align}
	\scriptsize
    \hat{C}^s = \sum_{k=1}^{S} f_k(\tilde{C}^s), s=1,2,...,S
\end{align}
where $f_k$ is a function to adaptively combine the cost volumes from different scales. We adopt the same definition of $f_k$ as HRNet \cite{hrnet2019}, which is defined as
\begin{align}
	\scriptsize
    f_k=
    \begin{cases}
    \mathcal{I}, k=s,\\
    (s-k)\text{ }3\times3\text{ convs with stride}=2, k < s, \\
    \text{unsampling }\oplus 1\times1\text{ conv}, k > s.
\end{cases}
\end{align}
where $\mathcal{I}$ denotes the identity function and $\oplus$ indicates bilinear up-sampling to the same resolution followed by a 1$\times$1 convolution to align the number of channels. In $f_k$, when $k<s$, $(s-k)$ 3$\times$3 convolutions with stride=2 are used for $2^{(s-k)}$ times down-sampling to make the resolution consistent.

In the cost aggregation module, we also support elastic scale which can be chosen from [2,3,4]. The chosen scale is naturally consistent with the scale number of feature pyramid.
Notice that for $S$ scales of cost volumes, the total number of combinations is $S^2/2$. Removing some scales can considerably reduce the computational efforts of cost aggregation. 

\textbf{Disparity Regression and Refinement.}
For each scale of the aggregated cost volumes, we use disparity regression as proposed in \cite{gcnet2017} to produce the estimated disparity map. The probability of each disparity $d$ is calculated from the predicted cost $C_{s}^{d}$ via the softmax operation $\sigma(\cdot)$. The estimated disparity $\hat{d}$ is calculated as the sum of each disparity candidate $d$ weighted by its probability.
\begin{align}
	\scriptsize
    \hat{d}=\sum_{d=0}^{D_{\text{max}}-1}d\times\sigma(c_d)
\end{align}
where $D_{\text{max}}$ is the maximum disparity range, $\sigma(\cdot)$ is the softmax function, and $c_d$ is the aggregated matching cost for disparity candidate $d$. As discussed in \cite{gcnet2017}, this regression has been proved to be more robust than using a convolution layer to directly produce the one-channel disparity map. In our EASNet, it will predict $S$ scales of disparity maps of different resolutions, from 1/3 to 1/(3$\times$2$^{S-1}$). 

Notice that the largest regressed disparity map is only 1/3 of the original resolution. To produce the full resolution of disparity, we apply the same two refinement modules in StereoDRNet \cite{stereodrnet2019} to hierarchically upsample and refine the predicted 1/3 disparity. The two refinement modules upsample the predicted disparity map from 1/3 to 1/2 and then 1/2 to full resolution, respectively. 

\subsection{Training EASNet}\label{subsec:train}
As discussed in \cite{cai2020once}, directly finetuning the network from scratch leads to prohibitive training cost and interference of model quality among different sub-networks. To efficiently train EASNet, we extend the progressive shrinking (PS) strategy of OFA to support our specialized search space of stereo matching. We first start with training the largest neural network (denoted as the full EASNet) with the maximum kernel size (K=7), depth (D=4), width (W=8) and scale (S=4). Then we perform four stages to finetune EASNet to support different dimensions of elastic factors. 

\textbf{Elastic Kernel Size, Depth and Width.} To search networks of different kernel sizes (K), depths (D) and widths (W), we apply the progressive shrinking strategy in \cite{cai2020once}, which is an effective and efficient training method to prevent interference among different
sub-networks. First, we support elastic kernel size which can choose from $\{3,5,7\}$ at each layer, while the depth and width remain the maximum values. This is achieved by  introducing kernel transformation matrices which share the kernel weights. For each layer, we have one $25\times25$ matrix and one $9\times9$ matrix that are shared among different channels, to transform the largest $7\times7$ kernels. Second, we support elastic depth. For a specific $D$, we keep the first $D$ layers and skip the last $(N-D)$ layers ($N$ is the original number of layers), which results that one depth setting only corresponds to one combination of layers. Third, we support elastic width. We introduce a channel sorting operation to support partial widths, which reorganizes the channels according to their importance (i.e., L1 norm of weights). Until now, we have finished three stages of model finetune. 

\textbf{Elastic Scale.} The scale search covers all four function modules in EASNet, and can be chosen from $\{2,3,4\}$. Take an example of choosing the scale $S$ for the largest scale $N$. Starting from feature pyramid construction, we keep the first $S$ scales of feature maps and skip the rest, which forms a $S$-level of feature pyramid. Then for the cost volume, $N-S$ cost volumes are naturally removed. Next for the cost aggregation, we only need to process $S^2/2$ combinations instead of $N^2/2$. Finally, the last $N-S$ scales of disparity maps are also skipped. In our experiments, this scale shrinking operation not only preserves the accuracy of larger sub-networks but also significantly reduces the network inference time.

%\textbf{Elastic Refinement.} Finally, our EASNet also supports elastic refinement, which is quantitatively defined as the number of refinement modules (R). As shown in Fig. \ref{fig:easnet}, our full EASNet has two disparity refinement modules (RM). Each RM takes the upsampled disparity as input and calculate its disparity residual map according to the error maps between the original left view and the warpped one from the right view. The predicted residual is then added to the input disparity map to improve its quality. In this finetune stage, we randomly skip the first RM and only pass the upsampled 1/2 disparity map to the second RM. In this way, we are able to obtain a smaller sub-network with satisfying prediction accuracy. 

\textbf{Loss Function.} Given a pair of rectified stereo RGB images, our EASNet takes them as inputs and produces $S+2$ disparity maps of different scales, where the first $S$ scales (denoted by $\hat{d_{s}^i}$) are generated by the AAModules and the rest two are generated by the refinement modules. We denote $\hat{d_{h}^i}$ as the first refinement result and $\hat{d_{f}^i}$ as the second one. Assume that the input image size is $H \times W$. For each predicted $\hat{d_i}$, it is first bilinearly upsampled to the full resolution. Then we adopt the pixel-wise smooth L1 loss to calculate the error between the predicted disparity map $\hat{d_i}$ and the ground truth $d_i$, 
\begin{align}
	\scriptsize
    L_s(d_s, \hat{d_s})=\frac{1}{N}\sum_{i=1}^{N}{smooth}_{L_1}(d_{s}^i - \hat{d_{s}^i}), s\in[1,...,S]\label{eq:smooth_l1}
\end{align}
where $N$ is the number of pixels of the disparity map, $d_s^i$ is the $i^{th}$ element of $d_s\in \mathcal{R}^N$ and 
\begin{align}
	\scriptsize
    {smooth}_{L_1}(x)=
    \begin{cases}
    0.5x^2,& \text{if } |x| < 1\\
    |x|-0.5,              & \text{otherwise}.
\end{cases}
\end{align}
The predicted refinement results $\hat{d_h}$ and $\hat{d_f}$ also follow the same smooth L1 loss calculation, denoted by $L_h$ and $L_f$ respectively. The final loss function is a weighted summation of losses over all disparity predictions as
\begin{equation}
	\scriptsize
    L=\sum_{s=1}^{S}w_sL_s(d_s,\hat{d_s})+w_hL_h(d_h,\hat{d_h})+w_fL_f(d_f,\hat{d_f}).\label{eq:weight_loss}
\end{equation}
In our experimental setting, the loss weights for the two lowest scale in \eqref{eq:weight_loss} are set to 1/3 and 2/3, while the rest are all set to 1.0.
\section{Evaluation}\label{sec:exp}
\subsection{Experimental Settings}
We conduct extensive experiments on four popular stereo datasets: Scene Flow \cite{mayer2016large}, MPI Sintel \cite{butler2012naturalistic}, KITTI 2012 \cite{kitti2012} and KITTI 2015 \cite{kitti2015}. 
%Scene Flow \cite{mayer2016large} is a large synthetic dataset which provides totally 39,824 samples of stereo RGB images. The width and height of the images are 960 and 540, respectively. The dataset covers a wide range of object shapes and texture and provides high-quality dense disparity ground truth. 
We use the training set of 35,454 samples of Scene Flow to train our EASNet, and then evaluate it on the test set of Scene Flow. 
%MPI Sintel \cite{butler2012naturalistic} has 1064 synthesized stereo images and ground truth data for disparity, of which the resolution is 1024$\times$436. It is derived from open-source 3D animated short film Sintel and has 23 different scenes. 
%The KITTI 2012 \cite{kitti2012} and KITTI 2015 \cite{kitti2015} are real-world datasets in the outdoor scenario, which only provide sparse ground truth. 
For Scene Flow and MPI Sintel, we use end-point error (EPE) to measure the accuracy of the methods, where EPE is the mean disparity error in pixels.
For KITTI 2012 and KITTI 2015, we report EPE and official metrics (e.g., D1-all) in the online leader board.

We implement our EASNet using PyTorch 1.8. First, we train the full network for 64 epochs with an initial learning rate of $1\times10^{-3}$. Then we follow the schedule described in Section \ref{subsec:train} to further fine-tune the full network, which contains five 25-epoch stages. The initial learning rate of each stage is set to $5\times10^{-4}$, which is decayed by half every 10 epochs. 

To compete for the methods in the online official leader board, we also fine-tune our EASNet on two KITTI datasets. We use a crop size of 336$\times$960, and first fine-tune the pre-trained Scene Flow model on mixed KITTI 2012 and 2015 training sets for 1000 epochs. The initial learning rate is $1\times10^{-3}$ and it is decreased by half every 300 epochs. Then another 1000 epochs are trained on the separate KITTI 2012/2015 training set for benchmarking, with an initial learning rate of $1\times10^{-4}$ and the same learning rate schedule as before.

As for data pre-processing, we follow the steps in \cite{xu2020aanet}, including color normalization and random cropping. 
%We perform color normalization with the mean ([0.485, 0.456, 0.406]) and variation ([0.229, 0.224, 0.225]) of the ImageNet \cite{deng2009imagenet} dataset for data pre-processing. During training, images are randomly cropped to size $H=384$ and $W=768$. 
For all the stages, we use the Adam (${\beta}_1=0.9, {\beta}_2=0.999$) optimizer to train EASNet. The network is trained with a batch size of 16 on 8 V100 GPUs. The entire architecture search optimization takes about 48 GPU days. Although AutoDispNet and LEAStereo only take 42 and 10 GPU days, the training cost of model re-searching and re-training can be prohibitive when they are applied to new computing devices.

To validate the deployment flexibility and efficiency, we benchmark our EASNet on three Nvidia GPUs with different computing levels, including the server-level Tesla V100, the desktop-level GTX 2070, and the inference-level Tesla P40.
\begin{figure*}[!t]
	\centering
	\includegraphics[width=0.96\linewidth]{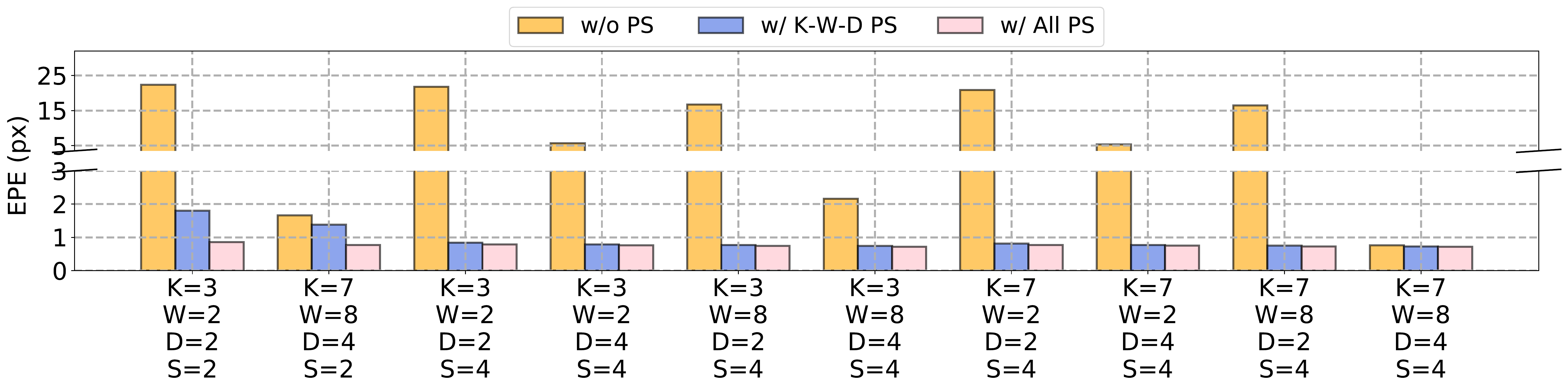}
	\caption{Scene Flow EPE (px) performance of sub-networks extracted from the full EASNet. $K$: kernel size, $W$: width, $D$: depth, $S$: Scale.}
	\label{fig:sf_comparison}
\end{figure*} 
\subsection{Experimental Results}
\textbf{Results of Different Sub-networks.} Fig. \ref{fig:sf_comparison} reports the Scene Flow EPEs of sub-networks derived from the full EASNet of different training schemes. Due to space limits, we take 10 sub-networks for comparison, and each of them is denoted as ``(K=$k$, W=$w$, D=$d$, S=$s$)''. It represents a sub-network that has $d$ layers for all units in the feature extraction module with the expansion ratio and kernel size set to $w$ and $k$ for all layers, and $s$ scales throughout all the function modules in EASNet. ``w/o PS'' indicates that we only train the largest full EASNet without model finetune, while ``w/ K-W-D PS'' and ``w/ All PS'' indicate that the full EASNet is fine-tuned using progressive shrinking (PS) of the first three stages (kernel size, width, and depth) and the complete five stages, respectively. Without PS, the model accuracy is significantly degraded while shrinking width and depth (seen from the last four sub-networks.). After performing PS for K-W-D, the accuracy of all the sub-networks can be improved by a significant margin. Moreover, our proposed shrinking scheme on scale (S) and refinement (R) can further reduce nearly 50\% of the estimation error (seen from the first two sub-networks). Specifically, without architecture optimization, our complete PS scheme can still achieve 0.86 of average EPE using only 0.78 M parameters under the architecture setting (K=3, W=2, D=2, S=2), which is on par with AANet (EPE: 0.87, 3.9 M parameters). In contrast, without the additional PS for scale and refinement, it only achieves 1.8, which is 0.94 worse.
\begin{figure*}[!t]
	\centering
	\includegraphics[width=0.9\linewidth]{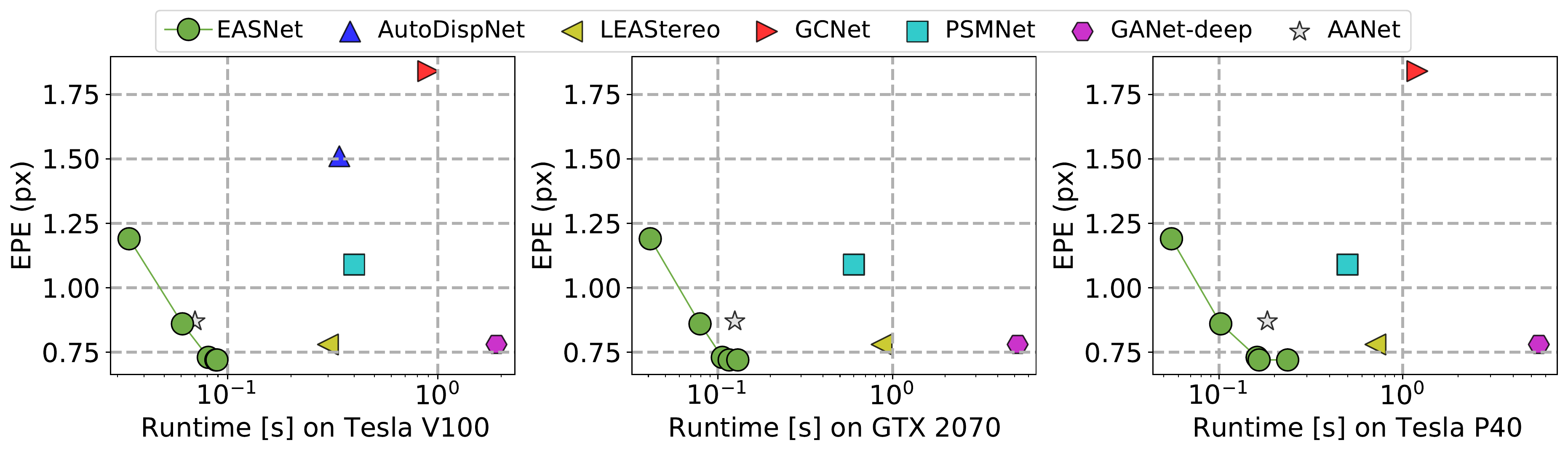}
	\caption{Our proposed method, EASNet, sets a new state-of-the-art on the Scene Flow \textsl{test} dataset with much fewer parameters and much lower inference time. The data points on the EASNet line indicate different sub-networks sampled from its largest full network structure.}
	\label{fig:sf_device}
\end{figure*} 

\begin{table}[!t]
	\centering
	\caption{Quantitative results on Scene Flow dataset. The runtime is measured on Nvidia Tesla P40. Bold indicates the best. Underline indicates the second best. Parentheses indicate that the results are reported by the original paper on Nvidia Tesla V100.}
	\label{tab:sf_results}
	\small{
		\begin{tabular}{l|c|c|c} \hline
			\textbf{Method} & \textbf{Params} [M] & \textbf{EPE} [px] & \textbf{Runtime} [s] \\ \hline
			%FADNet\cite{wang2020fadnet} & 144.3 & 0.83 & \textbf{0.07} \\
			PSMNet \cite{psmnet2018} & 5.22 & 1.09 & 0.5 \\
			GANet-deep \cite{ganet2019} & 6.58 & 0.78 & 5.5 \\
			AANet \cite{xu2020aanet} & 3.9 & 0.87 & 0.18 \\
			AutoDispNet-CSS \cite{Saikia_2019_ICCV} & 37 & 1.51 & (0.34) \\
			LEAStereo \cite{cheng2020hierarchical} & \underline{1.81} & 0.78 & 0.71 \\
			DeepPruner (best) \cite{deeppruner2019} & 7.1 & 0.86   &  0.18 \\ 
			DeepPruner (fast) \cite{deeppruner2019} & 7.1 & 0.97   &  \textbf{0.06} \\ \hline
			EASNet-L & 5.07 & \textbf{0.72} & 0.24 \\  
			EASNet-M & 3.03 & \underline{0.73} & 0.16 \\
			EASNet-S & \textbf{0.78} & 0.86 & \underline{0.10} \\ \hline
		\end{tabular}
	}
\end{table}
\begin{figure*}[!t]
	\captionsetup[subfigure]{labelformat=empty, farskip=0pt}
	\centering
	\rotatebox[origin=l]{90}{\scriptsize FT3D}
	\subfloat[]
	{
		\includegraphics[width=0.175\linewidth]{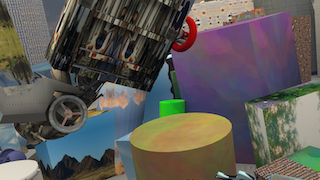}\label{fig:vt_bf_left_rgb}
	}
	\subfloat[]
	{
		\includegraphics[width=0.175\linewidth]{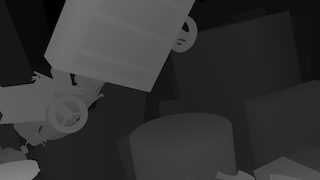}\label{fig:vt_bf_left_rgb}
	}
	\subfloat[]
	{
		\includegraphics[width=0.175\linewidth]{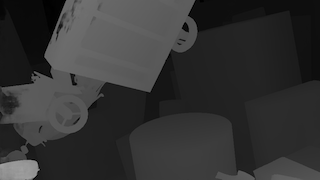}\label{fig:vt_bf_left_rgb}
	}
	\subfloat[]
	{
		\includegraphics[width=0.175\linewidth]{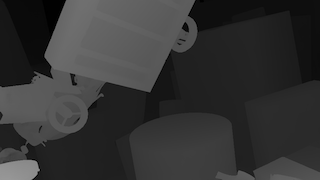}\label{fig:vt_bf_left_rgb}
	}
	\subfloat[]
	{
		\includegraphics[width=0.175\linewidth]{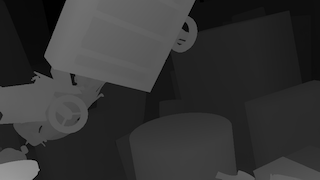}\label{fig:vt_bf_left_rgb}
	}
	\vspace{-1.0 em}
	\qquad
	\rotatebox[origin=l]{90}{\scriptsize Monkaa}
	\subfloat[]
	{
		\includegraphics[width=0.175\linewidth]{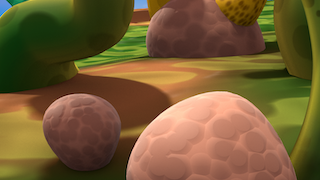}\label{fig:vt_bf_left_rgb}
	}
	\subfloat[]
	{
		\includegraphics[width=0.175\linewidth]{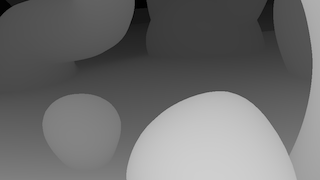}\label{fig:vt_bf_left_rgb}
	}
	\subfloat[]
	{
		\includegraphics[width=0.175\linewidth]{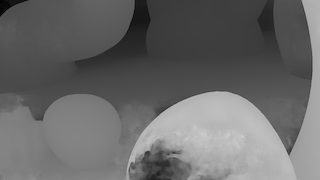}\label{fig:vt_bf_left_rgb}
	}
	\subfloat[]
	{
		\includegraphics[width=0.175\linewidth]{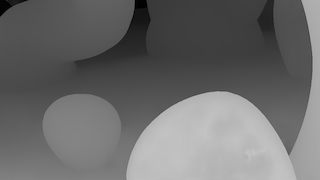}\label{fig:vt_bf_left_rgb}
	}
	\subfloat[]
	{
		\includegraphics[width=0.175\linewidth]{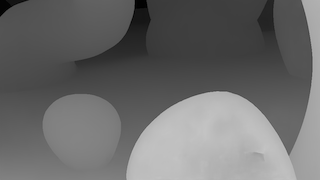}\label{fig:vt_bf_left_rgb}
	}
	\vspace{-1.0 em}
	\qquad
	\rotatebox[origin=l]{90}{\scriptsize Sintel}
	\subfloat[]
	{
		\includegraphics[width=0.175\linewidth]{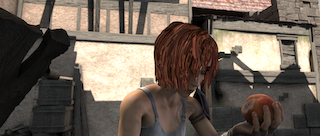}\label{fig:vt_bf_left_rgb}
	}
	\subfloat[\scriptsize GT]
	{
		\includegraphics[width=0.175\linewidth]{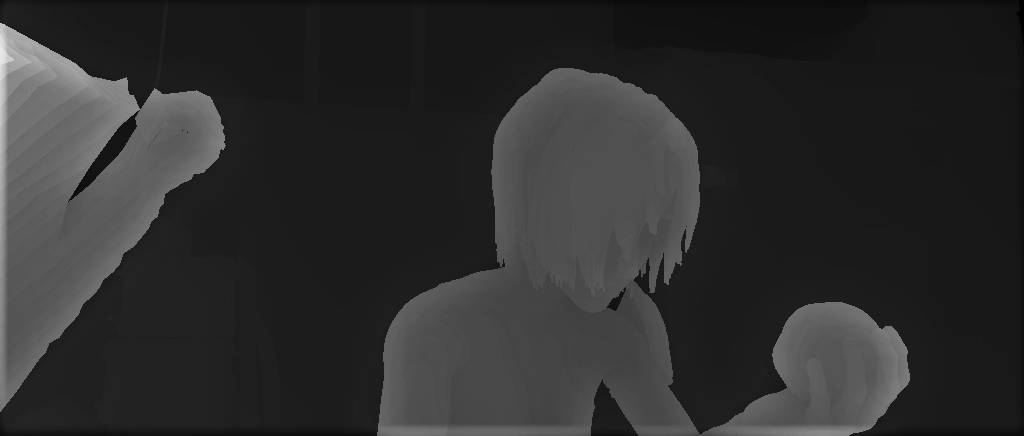}\label{fig:vt_bf_left_rgb}
	}
	\subfloat[\scriptsize AutoDispNet \cite{Saikia_2019_ICCV}]
	{
		\includegraphics[width=0.175\linewidth]{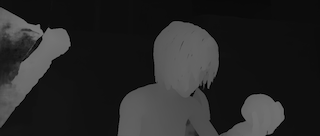}\label{fig:vt_bf_left_rgb}
	}
	\subfloat[\scriptsize LEAStereo \cite{cheng2020hierarchical}]
	{
		\includegraphics[width=0.175\linewidth]{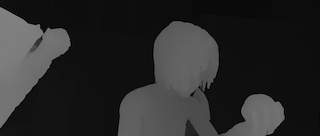}\label{fig:vt_bf_left_rgb}
	}
	\subfloat[\scriptsize EASNet]
	{
		\includegraphics[width=0.175\linewidth]{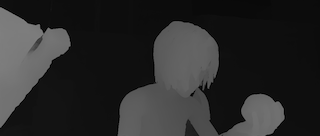}\label{fig:vt_bf_left_rgb}
	}
	\caption{Disparity predictions for the testing data of FlyingThings3D (FT3D), Monkaa and MPI Sintel. 
		%``FlyingThings3D'' and ``Monkaa'' are two subsets of Scene Flow. 
		The leftest column shows the left images of the stereo pairs. The rest four columns respectively show the disparity maps estimated by (a) ground truth, (b) AutoDispNet \cite{Saikia_2019_ICCV}, (c) LEAStereo \cite{cheng2020hierarchical}, and (d) our EASNet. 
		%EASNet achieves much better estimation quality than AutoDispNet on those pixels of the view boundary.
	}
	\label{fig:results_on_synthetic}
\end{figure*}

\textbf{EASNet under Different Hardware Computing Capability.} Fig. \ref{fig:sf_device} summarizes the results of different sub-networks extracted from EASNet under three GPUs. We also plot the results of other existing SOTA methods for comparison. First, EASNet outperforms all the other methods with Pareto optimality of both model accuracy and inference time. Take the desktop GPU GTX 2070 as an example. EASNet achieves a new SOTA 0.73 EPE with the runtime of 0.12 s, being 0.14 lower EPE than AANet that has similar inference performance. To achieve similar accuracy of AANet, EASNet performs 0.08 s on GTX 2070, which is 33.3\% lower than AANet. Second, since our EASNet only needs one time of training and does not need any further fine-tuning efforts when being deployed on devices of different computing capability, we can directly choose the sub-network from the full EASNet according to the latency requirement, while other methods cannot. For example, if we set the inference latency goal to 100 ms, for both the existing human-designed and NAS methods, only AANet on Tesla V100 can satisfy the requirement. However, our EASNet can provide a sub-network of competitive accuracy on all the three devices, i.e., 0.72 EPE with 0.09 s on Tesla V100, 0.73 EPE with 0.1 s on GTX 2070, and 0.86 EPE with 0.1 s on Tesla P40. This proves the flexibility and efficiency of our EASNet. 

\begin{table}[!t]
	\centering
	\caption{Quantitative results on other stereo datasets. Entries enclosed by parentheses indicate if they were tested on the target dataset without model finetuning. ``DN-CSS'' is short for DispNet-CSS. ``ADN-CSS'' is short for AutoDispNet-CSS. The time is measured on Nvidia Tesla V100 for KITTI resolution.}
	\addtolength{\tabcolsep}{-1.2pt}
	\label{tab:stereo_results}
	\small{
		\begin{tabular}{l||c|cccc|c} \hline
			\textbf{Method} & \textbf{Sintel} & \multicolumn{2}{c}{\textbf{KITTI}}  & \multicolumn{2}{c|}{\textbf{KITTI}} \\ 
			& (clean) & \multicolumn{2}{c}{(2012)}  & \multicolumn{2}{c|}{(2015)} & \textbf{Time} [s] \\ 
			& EPE & EPE & Out-noc & EPE & D1-all & \\
			& \textsl{train} & \textsl{train} & \textsl{test} & \textsl{train} & \textsl{test} & \\ \hline
			%FADNet\cite{wang2020fadnet} & 1.79 & 1.11 & - & 1.54 & - \\
			%FADNet-ft\cite{wang2020fadnet} & - & - & 2.04\% & - & 2.82\% \\
			ADN-CSS \cite{Saikia_2019_ICCV} & (2.14) & (\textbf{0.93}) & 1.70\% & (\textbf{1.14}) & 2.18\% & 0.34 \\
			GCNet \cite{gcnet2017} & - & - & 1.77\% & - & 2.87\% & 0.9 \\
			GANet \cite{ganet2019} & - & - & 1.19\% & - & 1.81\% & 1.9 \\
			AANet \cite{xu2020aanet} & - & - & 1.91\% & - & 2.55\% & 0.07 \\
			LEAStereo \cite{cheng2020hierarchical} & - & - & \textbf{1.13\%} & - & \textbf{1.65\%} & 0.3 \\ 
			DeepPruner (best) \cite{deeppruner2019} & - & - & - & - & 2.15\% & 0.18 \\\hline
			\multicolumn{7}{c}{$\leq$ 100 ms} \\ \hline
			DN-CSS \cite{flownet3} & (2.33) & (1.40) & 1.82\% & (1.37) & 2.19\% & 0.08 \\
			DeepPruner (fast) \cite{deeppruner2019} & - & - & - & - & 2.59\% & 0.06 \\
			MADNet \cite{madnet2019} & - & - & - & - & 4.66\% & \textbf{0.02} \\
			EASNet-L & (\textbf{1.58}) & (1.91) & 1.89\% & (1.90)  & 2.70\% & 0.09 \\
			EASNet-M & (1.59) & (1.91) &  1.96\% &  (2.18) & 2.89\% & 0.08 \\
			EASNet-S & (1.95) & (2.44) &  2.57\% &  (2.32) & 3.43\% & 0.06 \\ \hline
		\end{tabular}
	}
\end{table}
\begin{figure*}[!t]
	\captionsetup[subfigure]{labelformat=empty, farskip=0pt}
	\centering
	\rotatebox[origin=l]{90}{\scriptsize KITTI}
	\rotatebox[origin=l]{90}{\scriptsize 2012}
	\subfloat[]
	{
		\includegraphics[width=0.22\linewidth]{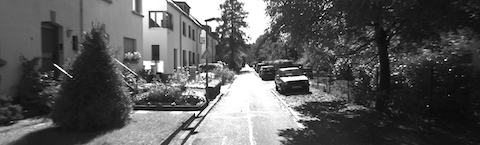}\label{fig:vt_bf_left_rgb}
	}
	\subfloat[]
	{
		\includegraphics[width=0.22\linewidth]{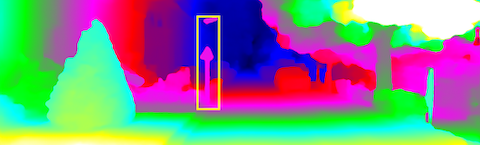}\label{fig:vt_bf_disp_gt}
	} % D1_all: 2.66\%
	\subfloat[]
	{
		\includegraphics[width=0.22\linewidth]{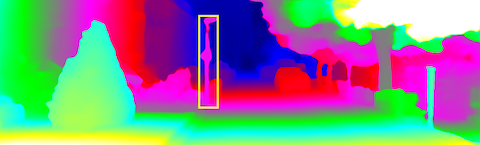}\label{fig:vt_bf_disp_gt} % D1_all: 1.35\%
	}
	\subfloat[]
	{
		\includegraphics[width=0.22\linewidth]{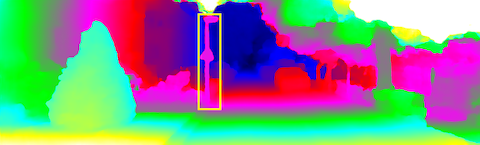}\label{fig:vt_bf_disp_gt} % D1_all: 2.02\%
	}
	\vspace{-1.0 em}
	\qquad
	\rotatebox[origin=l]{90}{\scriptsize KITTI}
	\rotatebox[origin=l]{90}{\scriptsize 2015}
	\subfloat[]
	{
		\includegraphics[width=0.22\linewidth]{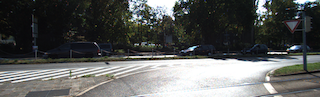}\label{fig:vt_bf_left_rgb}
	}
	\subfloat[\scriptsize AANet \cite{xu2020aanet}]
	{
		\includegraphics[width=0.22\linewidth]{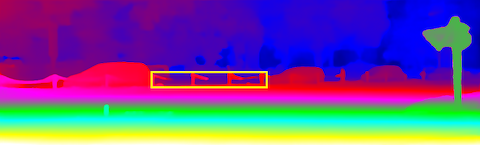}\label{fig:vt_bf_disp_gt}
	}
	\subfloat[\scriptsize LEAStereo \cite{cheng2020hierarchical}]
	{
		\includegraphics[width=0.22\linewidth]{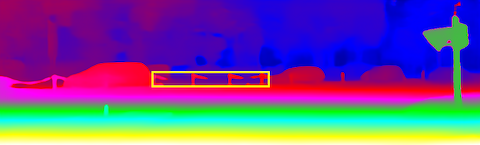}\label{fig:vt_bf_disp_gt}
	}
	\subfloat[\scriptsize EASNet]
	{
		\includegraphics[width=0.22\linewidth]{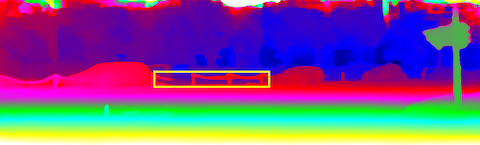}\label{fig:vt_bf_disp_gt}
	}
	\caption{Disparity predictions for KITTI 2012 and 2015 testing data. The leftest column shows the left images of the stereo pairs. The rest three columns show the disparity maps estimated by existing methods and our EASNet.}
	\label{fig:results_on_kitti}
\end{figure*}

\textbf{Benchmark Results on Scene Flow.} For the rest of experiments, we pick three sub-networks from the full EASNet, EASNet-L (K=$7$, W=$8$, D=$4$, S=$4$), EASNet-M (K=$7$, W=$8$, D=$2$, S=$4$), and EASNet-S (K=$3$, W=$2$, D=$2$, S=$2$).
%as listed in Table \ref{tab:eas_selected}. 
%A sub-network with $R=r$ indicates that it has $r$ refinement modules in Fig. \ref{fig:easnet}.

We compare our EASNet networks with five SOTA methods, including three hand-crafted and two NAS-based networks on Scene Flow \cite{mayer2016large} test set with 192 disparity level. In Table \ref{tab:sf_results}, we can observe that EASNet-M achieves the best performance using only near half of parameters in comparison to the SOTA hand-crafted methods (e.g., GANet \cite{ganet2019}). Furthermore, the previous SOTA NAS-based method AutoDispNet \cite{Saikia_2019_ICCV} has 10$\times$ more parameters than our EASNet-M. Our smallest sub-network EASNet-S can still achieve much better accuracy than AutoDispNet and comparable accuracy to AANet with much fewer parameters and faster inference speed. As for the model runtime, EASNet-L outperforms the accuracy SOTA, GANet \cite{ganet2019} and LEAStereo \cite{cheng2020hierarchical} by 3$\times$ and 22$\times$ respectively. EASNet-M achieves Pareto optimality in both accuracy and speed among all the methods. We show some of the qualitative results in Fig. \ref{fig:results_on_synthetic}. Our EASNet outperforms AutoDispNet in terms of estimation quality and achieves competitive accuracy to LEAStereo with only one third of inference time on P40.

\iffalse
\begin{table}[!t]
	\centering
	\caption{Selected Sub-networks from EASNet.}
	\label{tab:eas_selected}
	\small{
	\begin{tabular}{c|c|c|c|c|c} \hline
		\textbf{Name} & \textbf{D} & \textbf{W} & \textbf{K} & \textbf{S} & \textbf{R} \\ \hline
		EASNet-L & 4 & 8 & 7 & 4 & 2 \\ \hline
		EASNet-M & 2 & 8 & 7 & 4 & 2 \\ \hline
		EASNet-S & 2 & 2 & 3 & 2 & 2 \\ \hline
	\end{tabular}
	}
\end{table}
\fi

\textbf{Benchmark Results on Sintel and KITTI.} We evaluate the model generalization of EASNet on the other two datasets, MPI Sintel and KITTI. Table \ref{tab:stereo_results} shows the results. Sintel is tested without any model finetune. EASNet-L achieves a much lower EPE than DispNet-CSS and AutoDispNet-CSS. 
%Notice the models finetuned on the KITTI datasets (e.g., the entries of DispNet-CSS-ft and AutoDispNet-CSS-ft) will critically lose generalization. We do not expect better results than our EASNet-L on Sintel from the rest ``-ft'' methods. 
Besides, after being finetuned on the KITTI training data, EASNet still shows satisfying accuracy with the state of the art on the common public benchmarks. EASNet-L achieves the best or second best accuracy among the methods of less than 100 ms. EASNet-S also achieves competitive accuracy with much lower latency. We show some of the qualitative results in Fig. \ref{fig:results_on_kitti}. Our EASNet is able to capture the disparity information of those thin objects, such as street light and road fence.

\subsection{Discussion}
There are several hints from the experiments. 1. The scale of the feature extraction module does not need to be large to achieve a good performance, due to the fact that the sub-network of (S=2) has similar accuracy to that of (S=4) with our training strategy; 2. The inference time drop of EASNet-S mainly comes from shrinking the feature extraction units and the whole scale (nearly 58\%); 
%3. It is observed that skipping the first refinement module can further improve the inference speed of EASNet-S by 40\% but increase the EPE from 0.86 to 1.19. 
%Notice that we yet do not explore network search techniques for the refinement module. This indicates that our EASNet still has great potentials for deriving smaller sub-networks with the consistent accuracy.
\section{Conclusion and Future Work}\label{sec:conclusion}
In this paper, we proposed EASNet, an elastic and accurate stereo matching network that leverages the domain knowledge of the CVM-Conv3D methods to design a specialized search space covering enormous architecture settings. 
To efficiently train EASNet with the target of maximizing the accuracy of all the sub-networks, we use the progressive shrinking strategy to support the specialized network search space of four dimensions,
%derived from the domain knowledge of stereo matching, 
including depth, width, kernel size, and scale.
Superior to the previous studies that design and train a neural network for each deployment scenario, our EASNet can quickly generate the sub-networks that satisfy the deployment requirement of accuracy and latency.
Validated on public benchmarks among three devices of different computing capability, our EASNet achieves Pareto optimality in terms of model accuracy and inference speed among all state-of-the-art CVA-3D deep stereo matching architectures (human designed and NAS searched). 
%Given the target deployment device and the latency constraint, a sufficiently accurate sub-network can be directly extracted from the full EASNet without any additional model training. 

In the future, we plan to apply NAS to search the units of cost aggregation and disparity refinement, which owns great potential for deriving smaller sub-networks with consistent accuracy. Furthermore, how to combine the network search strategies of DARTS (for operator and cell link) and OFA (for cell/layer/block hyper-parameter) is also an interesting and potential direction of searching an efficient and effective network structure for stereo matching.

\section*{Acknowledgements}
This work was supported in part by the Hong Kong RGC GRF grant under the contract HKBU 12200418, grant RMGS2019\_1\_23 and grant RMGS21EG01 from Hong Kong Research Matching Grant Scheme, the NVIDIA Academic Hardware Grant, and Guangdong Provincial Key Laboratory of Novel Security Intelligence Technologies (2022B1212010005). 

\bibliographystyle{splncs04}
%\bibliography{stereo_matching}
\bibliography{main.bbl}
\end{document}